\def\year{2019}\relax
\documentclass[letterpaper]{article} 
\usepackage{aaai19}  
\usepackage{times}  
\usepackage{helvet} 
\usepackage{courier}  
\usepackage[hyphens]{url}  
\usepackage{graphicx} 
\usepackage{graphicx}  
\usepackage{amssymb}
\usepackage{amsthm}
\usepackage{amsmath}
\usepackage{comment}
\usepackage{algorithm}
\usepackage{algorithmic}
\usepackage{moresize}
\usepackage{helvet}
\usepackage{courier}
\usepackage{wrapfig}
\usepackage[font=small]{caption}
\usepackage{subcaption}
\def\i#1{\hbox{\it #1\/}}
\def\beq{\begin{equation}}
\def\eeq#1{\label{#1}\end{equation}}
\def\ba{\begin{array}}
\def\ea{\end{array}}

\def\t{\textbf{t}}
\def\f{\textbf{f}}

\def\iif{\hbox{\bf if}}

\def\causes{\hbox{\bf causes}}

\title{A Joint Planning and Learning Framework for Human-Aided Decision-Making}

\author{
Daoming Lyu$^1$, 
Fangkai Yang$^2$, 
Steven Gustafson$^3$,
Bo Liu$^1$ \thanks{Correspondence to: Bo Liu$<$\texttt{boliu@auburn.edu}$>$.}
\\ 
$^1$ Auburn University, Auburn, AL, USA\\
$^2$ NVIDIA Corporation, Redmond, WA, USA\\
$^3$ Maana Inc., Bellevue, WA, USA  \\
daoming.lyu@auburn.edu,
fangkaiy@nvidia.com,
steven.gustafson@gmail.com,
boliu@auburn.edu
}

\begin{document}

\maketitle

\begin{abstract}

	Conventional reinforcement learning (RL) allows an agent to learn policies via environmental rewards only, with a long and slow learning curve, especially at the beginning stage. On the contrary, human learning is usually much faster because prior and general knowledge and multiple information resources are utilized. In this paper, we propose a \textbf{P}lanner-\textbf{A}ctor-\textbf{C}ritic architecture for hu\textbf{MAN}-centered planning and learning (\textbf{PACMAN}), where an agent uses prior, high-level, deterministic symbolic knowledge to plan for goal-directed actions. PACMAN integrates Actor-Critic algorithm of RL to fine-tune its behavior towards both environmental rewards and human feedback. To the best our knowledge, This is the first unified framework where knowledge-based planning, RL, and human teaching jointly contribute to the policy learning of an agent. Our experiments demonstrate that PACMAN leads to a significant jump-start at the early stage of learning, converges rapidly and with small variance, and is robust to inconsistent, infrequent, and misleading feedback.
\end{abstract}

\section{Introduction}

A longstanding goal of artificial intelligence is to enable the programming agent to perform tasks intelligently in a complex domain. Recently, reinforcement learning (RL) algorithms, such as DQN, have made a lot of success on training agent to play Atari games from raw pixel images~\cite{dqn:nature:2015}. However, this approach is criticized for being ``data-hungry'' and ``time-hungry'', although it can learn fine granular policies that surpass human experts. Such drawbacks are heavily related to two phenomena that have not drawn enough attention before. 
The first phenomenon is that conventional RL algorithms can only learn from the reward signal, thus limiting its capability of utilizing multiple information resources.
In general, humans can learn from multiple resources, and supervised learning that can learn from multiple labels simultaneously in the vector space. On the contrary, RL algorithms can only learn from scalar reward signals by interacting with the environment, i.e., one scalar reward signal per iteration. It would be beneficial to learn from multiple resources beyond merely environmental rewards, such as human feedback.
The second phenomenon involves the initial learning phase.  Existing RL algorithms usually learn from scratch, and thus often lead to an inferior performance at the initial learning phase~\cite{sutton2018reinforcement}, with a very long and slow learning curve. It would be helpful if a certain amount of prior knowledge can be incorporated in advance to help improve the initial learning dynamics.

There have been some studies, along with the two aforementioned research directions. The first topic leads to the proposal of ``human-centered reinforcement learning" (HCRL) where an agent learns directly from human feedback. These approaches include interpreting human feedback as a shaping reward~\cite{knox2009interactively}, applying human feedback directly to policy improvement \cite{thomaz2008teachable,knox2010combining,griffith2013policy}, or interpreting human feedback as an estimation of the advantage function $A{^\pi}(s,a)$ \footnote{An advantage function $A{^\pi}(s,a)$ is a state-action value roughly corresponding to how much better or worse an action $a$ is compared to the current
	policy at state $s$.}~\cite{macglashan2016convergent,macglashan2017interactive}. However, all the aforementioned work is limited to learning human feedback only, without considering the environmental rewards. 
The second topic, learning from human's prior knowledge, is investigated within a rather limited scope. To the best of our knowledge, the most relevant research to this topic is learning from demonstrations (LfD)~\cite{lfd:argall:2009}, where the original policy search problem is reduced to a supervised distribution matching problem by matching the expert's demonstration trajectory's distribution. A typical strategy is to apply LfD first to obtain a good initial policy and then use RL methods for further policy improvement and refinement.
The alternative approach, \textit{learning from explicitly represented, formal, symbolic knowledge} is neglected until recently. 
Symbolic knowledge is used to capture coarse-granular domain dynamics and provide general guidance to exploration, leading to jump-start at the early stage of learning. After that,  RL is used to fine-tune further the performance~\cite{leonetti2016synthesis,yang2018peorl,lyu2018sdrl,yang:iros:2019}, which can significantly improve the sample-efficiency. 

In this paper, we argue that prior knowledge, learning from environmental rewards, and human teaching should jointly contribute to obtaining the optimal behavior. By representing at a sufficiently abstract level rather than specifically tailored towards individual problems, symbolic knowledge can be light-weight and concise to be a useful guideline for data-driven learning. The agent can further learn domain details and uncertainties to refine its behavior simultaneously from both environmental rewards and human feedback. In this way, the prior knowledge is enriched with experience and tailored towards individual problem instances. Based on the motivation above, we propose the \textbf{P}lanner--\textbf{A}ctor--\textbf{C}ritic architecture for hu\textbf{MAN} centered planning and learning (\textbf{PACMAN}). PACMAN interprets human feedback as the advantage function estimation similar to COACH framework but further incorporates prior, symbolic knowledge. The contributions of this paper are summarized as follows:
\textbf{(i.)} The framework of PACMAN features symbolic planner-actor-critic trio iteration, where planning and RL mutually benefit each other. In particular, the logical representation of action effects is dynamically generated by sampling a stochastic policy learned from actor-critic (AC) algorithm of RL. PACMAN allows the symbolic knowledge and actor-critic framework to integrate into a unified framework seamlessly.
\textbf{(ii.)} This framework enables joint learning from both environmental rewards and human feedback, which can accelerate the learning process of interactive RL as well as improve the tolerance of misleading feedback from human users. 

To the best of our knowledge, this paper is the first work that learns simultaneously from human feedback, environmental rewards, and prior symbolic knowledge. While our framework can be quite generic, we choose to use ASP-based action language $\mathcal{BC}$ \cite{lee13}, answer set solver {\sc clingo} to perform symbolic planning and conduct our experiments. 
The evaluation of the framework is performed on RL benchmark problems such as Four Rooms and Taxi domains. We consider various scenarios of human feedback, including the cases of ideal, infrequent, inconsistent, and both infrequent and inconsistent with helpful feedback and misleading feedback, and compare our approach with the state-of-the-art methods. Our experiments indicate that PACMAN empirically leads to a significant jump-start at early stages of learning, converges faster and with smaller variance, and is robust to inconsistent, infrequent cases even misleading feedback.

\section{Related Work}
\label{sec:relatedwork}
There is a long history of work that combines symbolic planning with reinforcement learning \cite{parr1998reinforcement,Ryan98rl-tops:an,Ryan02usingabstract,hogg2010learning,leonetti2012automatic,leonetti2016synthesis}. These approaches were based on integrating symbolic planning with value iteration methods, and thus planning and learning cannot be mutually beneficial to each other. 
The latest work in this direction is PEORL framework \cite{yang2018peorl} and SDRL \cite{lyu2018sdrl}, where ASP-based planning was integrated with R-learning \cite{rlearning:schwartz} into planning--learning loop. PACMAN architecture is a new framework of integrating symbolic planning with RL, in particular, integrating planning with AC algorithm for the first time, and also features bidirectional communication between planning and learning.

Learning from human feedback takes the framework of reinforcement learning, and incorporate human feedback into reward structure\cite{thomaz2006reinforcement,knox2009interactively,knox2012reinforcement}, information directly on policy \cite{thomaz2008teachable,knox2010combining,griffith2013policy}, or advantage function \cite{macglashan2016convergent,macglashan2017interactive}.
Learning from both human feedback and environmental rewards were investigated \cite{thomaz2006reinforcement,knox2012reinforcement,griffith2013policy}, mainly integrating the human feedback to reward or value function via reward shaping or Q-value shaping. 
Such methods do not handle well the samples with missing human feedback, and in reality, human feedback may be infrequent. 

Recent work of COACH \cite{macglashan2016convergent,macglashan2017interactive} showed that human feedback seems better to formulate as an estimation of the policy-dependent advantage function, but it does not consider learning simultaneously from environmental rewards and human feedback. Besides, none of these work considers the setting where an agent is equipped with prior knowledge and generates a goal-directed plan that is further to be fine-tuned by reinforcement learning and a human user. By integrating human feedback into PACMAN, our framework allows the integration of logic-based symbolic planning into the data-driven learning process, where environmental rewards and human feedback can be unified into advantage function to shape the agent's behavior in the context of long-term planning. 

\section{Preliminaries}
\label{sec:prelim}

\paragraph{Symbolic Planning}
Symbolic planning concerns on describing preconditions and effects of actions using a formal language and automated plan generation, which has been used for high-level task planning in a variety of robotic applications \cite{hanheide2015robot,khandelwal2017bwibots}. An {\em action description} $D$ in the language $\mathcal{BC}$ includes two kinds of symbols, {\em fluent constants} that represent the properties of the world, with the signature denoted as $\sigma_F(D)$, and {\em action constants}, with the signature denoted as $\sigma_A(D)$. A \textit{fluent atom} is an expression of the form $f=v$, where $f$ is a fluent constant and $v$ is an element of its domain. For Boolean domain, denote $f=\t$ as $f$ and~$f=\f$ as $\sim\!\!\! f$. An action description is a finite set of {\em causal laws} that describe how fluent atoms are related with each other in a single time step, or how their values are changed from one step to another, possibly by executing actions. For instance,
$
A~\iif~A_1,\ldots,A_m
$
is a {\em static law} that states at a time step, if $A_1,\ldots, A_m$ holds then $A$ is true. 
$
a~\causes~A_0~\iif~A_1,\ldots, A_m
$
is a {\em dynamic law}, stating that at any time step, if $A_1,\ldots, A_m$ holds, by executing action $a$,~$A_0$ holds in the next step.\footnote{In $\mathcal{BC}$, causal laws are defined in a more general form. In this paper, without loss of generality, we assume the above form of causal laws for defining effects of actions.} An action description captures a dynamic transition system. 
Let $I$ and $G$ be states. The triple $(I,G,D)$ is called a planning problem. $(I,G,D)$ has a plan of length $l-1$ \textbf{iff} there exists a transition path of length $l$ such that $I=s_1$ and $G=s_l$. Throughout the paper, we use $\Pi$ to denote both the plan and the transition path by following the plan. Generating a plan of length $l$ can be achieved by solving the answer set program $\i{PN}_l(D)$, consisting of rules translated from $D$ and appending timestamps from 1 to $l$, via a translating function $\i{PN}$. For instance, $\i{PN}_l$ turns the static law to
$
i\!:\!A\leftarrow i\!:\!A_1,\ldots, i\!:\!A_m,
$
where $1\le i\le l$
and turns the dynamic law to 
$
i+1\!:\!A\leftarrow i\!:\!a, i\!:\!A_1,\ldots, i\!:\!A_m,
$
where $1\le i < l$. See \cite{lee13} for details.

\paragraph{Reinforcement Learning and Actor-Critic Method}


RL problem is usually defined as a Markov Decision Process (MDP), which is a tuple of $({\mathcal{S},\mathcal{A},P_{ss'}^{a},r,\gamma})$. Specifically, $\mathcal{S}$ and $\mathcal{A}$ denotes state space and action space, the transition kernel $P_{ss'}^{a}$ specifies the probability of transition from state $s\in\mathcal{S}$ to state $s'\in\mathcal{S}$ by taking action $a\in\mathcal{A}$, $r(s,a):\mathcal{S}\times\mathcal{A}\mapsto\mathbb{R}$ is a reward function bounded by $r_{\max}$, and $0\leq\gamma<1$ is a discount factor. RL concerns learning a near-optimal policy $\pi(a|s)$ (maps a state $s$ to an action $a$) by executing actions and observing the state transitions and rewards.

An actor-critic~\cite{nac:peters2008,nac:Bhatnagar09,sutton2018reinforcement} approach is a framework of RL which has two components: the actor and the critic. Typically, the actor is a policy function $\pi_{\theta}(a|s)$ parameterized by $\theta$ for action selection, while the critic is a state-value function $V_{x}(s)$ parameterized by $x$ to criticize the action made by the actor. For example, after each action selection, the critic will evaluate the new state to determine whether things have gone better or worse than expected by computing TD error \cite{sutton2018reinforcement}.
If the TD error is positive, it suggests that the tendency to select current action $a$ should be strengthened for the future, whereas if the TD error is negative, it suggests the tendency should be weakened. This TD error is actually an unbiased estimation of advantage function $A{^\pi}(s,a)$ \cite{schulman2015high}.



\section{PACMAN Architecture}
\label{sec:pacman_arch}
\begin{figure}[!tbp]
\centering
\includegraphics[height=4.2cm,width=6.3cm]{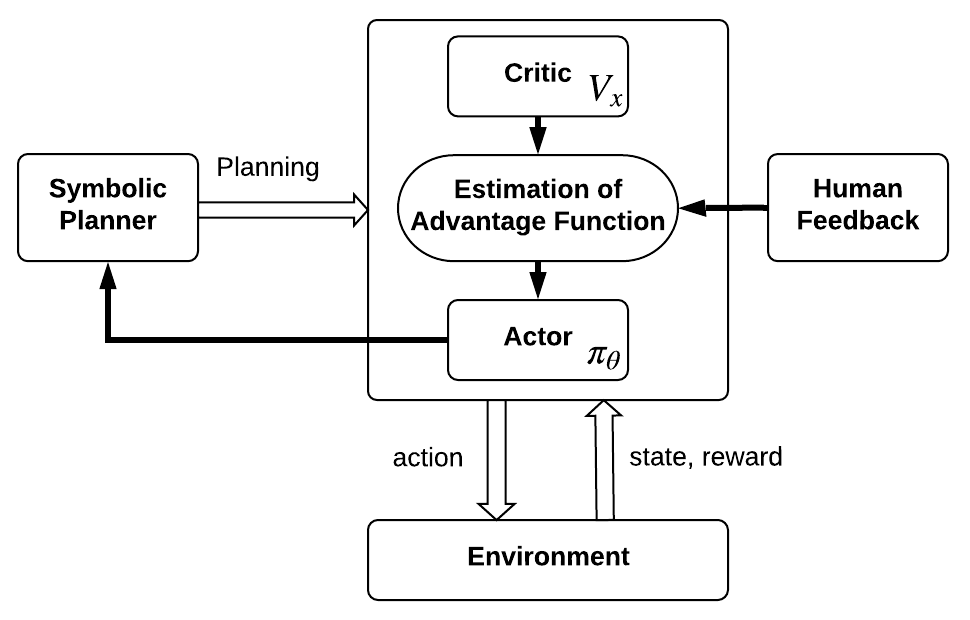}
\caption{Architecture illustration}
\label{fig:arch}
\end{figure}

In this section, we will present our PACMAN architecture, which is shown in Figure~\ref{fig:arch}. With the encoded prior knowledge and the policy function (from the actor), the symbolic planner would generate a plan that contains a sequence of actions, and send it to RL (actor-critic) to execute. During the interaction between RL and environment, the estimation of advantage function can be either from TD error computed by the critic or the value of human feedback. The detailed process will be defined formally as follows.

\subsection{Sample-based Symbolic Planning}
\begin{figure}[!tbp]
\centering
\includegraphics[width=0.49\textwidth]{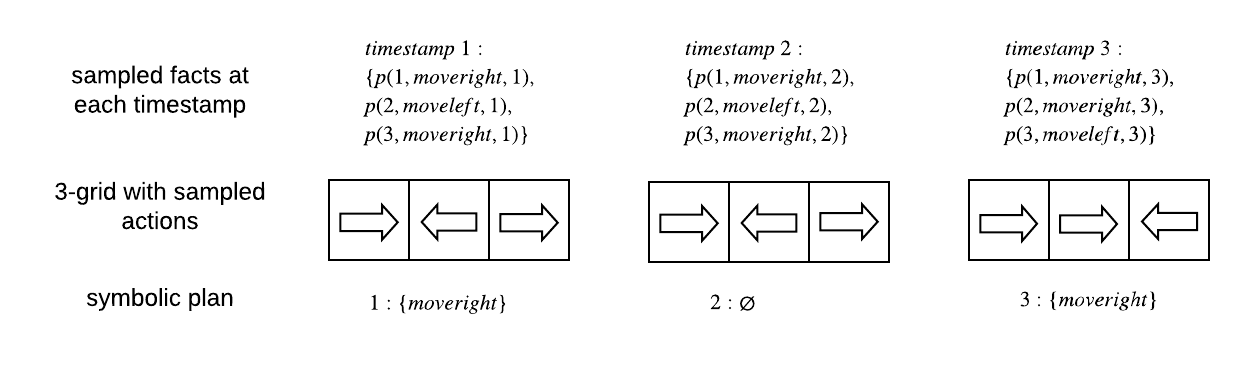}
\caption{A possible sample-based planning result for 3-grid domain}
\label{fig:sample}
\end{figure}

We introduce a {\em sample-based planning problem} as a tuple $(I,G,D,\pi_{\theta})$ where $I$ is the initial state condition, $G$ is a goal state condition, $D$ is an action description in $\mathcal{BC}$, and~$\pi_{\theta}$ is a stochastic policy function parameterized by~$\theta$, i.e., a mapping $\mathcal{S}\times \mathcal{A} \mapsto [0,1]$. For $D$, defines its $l$-step {\em sampled action description} $D^l_{\pi} = D_s\cup D_d\cup \bigcup_{t=1}^l P^t_{\pi}$ with respect to policy $\pi$ and time stamp $1\le t\le l$, where
\begin{itemize}
\item $D_s$ is a set of causal laws consisting of static laws and dynamic laws that does not contains action symbols;
\item $D_d$ is a set of causal laws obtained by turning each dynamic law of the form
$
a~\causes~A_0~\iif~A_1,\ldots,A_m,
$
into rules of the form
$
a~\causes~A_0~\iif~A_1,\ldots,A_m,p(s,a)
$
where $\i{p}$ is a newly introduced fluent symbol and $\{A_1,\ldots, A_m\}\subseteq s$, for $s\in\mathcal{S}$; and
\item $P^t_{\pi}$ is a set of facts sampled at timestamp $t$ that contains $p(a,s)$ such that
$
p(s,a)\in P^t_{\pi}\sim \pi(\cdot|s,\theta)
$
where for $s\in\mathcal{S}$, $A\in\mathcal{A}$.
\end{itemize}
Define translation $\mathcal{T}(D^l_{\pi})$ as
$
\i{PN}_l(D_s\cup D_d)\cup\bigcup_{t=1}^l\{p(s,a,t), \hbox{for}~p(s,a)\in P^t_\pi\}
$
that turns $D^l_\pi$ into answer set program.
A {\em sample-based plan} up to length $l$ of $(I,G,D,\pi_{\theta})$ can be calculated from the answer set of program $\mathcal{T}(D^l_{\pi})$ such that $I$ and $G$ are satisfied. The planning algorithm is shown in Algorithm~\ref{alg:splanning}.

\begin{algorithm}[!tbp]
  \caption{Sample-based Symbolic Planning}
  \label{alg:splanning}
  \begin{algorithmic}[1]
    \REQUIRE a sample based planning problem $(I,G,D,\pi_{\theta})$
    \STATE $\Pi\Leftarrow\emptyset$, calculate $D_{\pi}^0$,  $k\Leftarrow 1$
      \WHILE{$\Pi=\emptyset$ and $k<\i{maxstamp}$}    
        \STATE sample $P_{\pi}^k$ over $p(s,a)\sim\pi_{\theta}(\cdot|s)$ for $s\in\mathcal{S}$, $a\in\mathcal{A}$
        \STATE $D^k_{\pi} \Leftarrow D^{k-1}_{\pi}\cup P^k_{\pi}$
        \STATE $\displaystyle\Pi\leftarrow\textsc{Clingo}.\i{solve}(I\cup G\cup {\mathcal{T}}(D^k_{\pi}))$
        \STATE $k\leftarrow k+1$
      \ENDWHILE
   \RETURN $\Pi$
  \end{algorithmic}
\end{algorithm}

\noindent\textbf{Example.} Consider 3$\times$1 horizontal gridworld where the grids are marked as state 1, 2, 3, horizontally. Initially the agent is located in state 1. The goal is to be located in state 3. The agent can move to left or right. Using action language ${\mathcal{BC}}$, moving to the left and moving to the right can be formulated as dynamic laws
$$
\ba{rl}
\!\!\!\i{moveleft}&\!\!\!\causes~\i{Loc}=L-1~\iif~\i{Loc}=L.\\
\!\!\!\i{moveright}&\!\!\!\causes~\i{Loc}=L+1~\iif~\i{Loc}=L.
\ea
$$
Turning them into sample-based action description leads to
$$
\ba{rlr}
\i{moveleft}&\!\!\!\causes~\i{Loc}=L-1~\iif~\i{Loc}=L,p(L,\i{moveleft}).\\
\i{moveright}&\!\!\!\causes~\i{Loc}=L+1~\iif~\i{Loc}=L,p(L,\i{moveright}).
\ea
$$
The policy estimator $\pi_{\theta}$ accepts an input state and output probability distribution on actions $\i{moveleft}$ and $\i{moveright}$. Sampling $\pi_{\theta}$ with input $s$ at time stamp $i$ generates a fact of the form $p(s,a,i)$ where $a\in\{\i{moveleft},\i{moveright}\}$ following the probability distribution of $\pi_{\theta}(\cdot|s)$.

At any timestamp, \textsc{clingo} solves answer set program consisting of rules translated from the above causal laws:
{\scriptsize
\begin{verbatim}
loc(L-1,k+1):-moveleft(k),loc(L,k), p(L,moveleft,k).
loc(L+1,k+1):-moveright(k),loc(L,k), p(L,moveright,k).
\end{verbatim}}
\noindent for time stamp $1,\ldots,k$, plus a set of facts of the form {\tt p(s,a,i)} sampled from $\pi_{\theta}$ where for states $s\in\{1,2,3\}$ and timestamps $i\in\{1,\ldots,k\}$. Note that the planner can skip time stamps if there is no possible actions to use to generate plan, based on sampled results. Figure~\ref{fig:sample} shows a possible sampling results over 3 timestamps, and a plan of 2 steps is generated to achieve the goal, where time stamp 2 is skipped with no planned actions.
Since sample-based planning calls a policy approximator as an oracle to obtain probability distribution and samples the distribution to obtain available actions, it can be easily applied to other planning techniques such as PDDL planning. For instance, the policy appropriator can be used along with heuristics on relaxed planning graph \cite{helmert2006fast}.

\subsection{Planning and Learning Loop}

The planning and learning loop for PACMAN, as shown in Algorithm~\ref{alg:pac}, starts from a random policy (uniform distribution over action space), and then generate a sample-based symbolic plan. After that, it follows the plan to explore and update the policy function $\pi_{\theta}$, leading to an improved policy, which is used to generate the next plan.

\begin{algorithm}
\caption{PACMAN}
\label{alg:pac}
\begin{algorithmic}[1]
\REQUIRE $(I,G,D,\pi_{\theta})$ and a value function estimator $V_x$
\FOR {$\i{episode}=0,1,\dots,maxepisode$}
\STATE Generate symbolic plan $\Pi$ from $(I,G,D,\pi_{\theta})$ by Algorithm~\ref{alg:splanning}
\FOR {$\langle s_{i}, a_{i}, r_{i}, s_{i+1} \rangle\in\Pi$}
\STATE Compute TD error as ${\delta _i} = {r_i} + \gamma {V_{{x_{i+1}}}}({s_{i + 1}}) - {V_{{x_{i+1}}}}({s_i})$.
\STATE Update $V_x$ via $
{x_{i + 1}} = {x_i} + \alpha {\delta _i}\nabla {V_{{x_i}}}({s_i})
$. 
\IF{human feedback $f_i$ is available}
\STATE Replace TD error ${\delta _i}$ with human feedback $f_i$.
\ENDIF
\STATE Update $\pi_{\theta}$ via $
\theta_{i+1} = \theta_{i} + \beta {\delta _i}\nabla \log \pi_{\theta}(a_i|s_i)
$. 
\ENDFOR
\ENDFOR
\end{algorithmic}
\end{algorithm}

For the $i$-th experience tuple of an episode, $(s_{i}, a_{i}, r_{i}, s_{i+1})$, the TD error is computed as
$
{\delta _i} = {r_i} + \gamma {V_{{x_{i+1}}}}({s_{i + 1}}) - {V_{{x_{i+1}}}}({s_i}),
$
which is a stochastic estimation of the advantage function. 
The value function $V_{x}$ is updated using reinforcement learning approaches, such as TD method~\cite{sutton2018reinforcement}:
$
{x_{i + 1}} = {x_i} + \alpha {\delta _i}\nabla {V_{{x_i}}}({s_i}),
$
where $\alpha$ is the learning rate.
The policy function $\pi_{\theta}$ will be updated by
$
\theta_{i+1} = \theta_{i} + \beta {\delta _i}\nabla \log \pi_{\theta}(a_i|s_i),
$
where $\beta$ is the learning rate.
If the human feedback signal $f_i$ is available, then this feedback signal will replace the previous computed TD error and be used to update the policy function; If there is no human feedback signal available at this iteration, TD error will be used to update the policy function directly. For this reason, human feedback here can be interpreted as guiding exploration towards human preferred state-action pairs.


\section{Experiment}
\label{sec:exp_and_res}

We evaluate our method in two RL-benchmark problems: Four Rooms \cite{sutton1999between} and Taxi domain \cite{barto-sm:hrl}. For experiments, we consider the discrete value of (positive or negative) feedback with the cases of ideal (feedback is always available without reverting), infrequent (only giving feedback at 50\% probability), inconsistent (randomly reverting feedback at 30\% probability) and infrequent+inconsistent (only giving feedback at 50\% probability, while randomly reverting feedback at 30\% probability). We compare the performance of PACMAN with 3 methods: TAMER+RL Reward Shaping from \cite{knox2012reinforcement}, BQL Reward Shaping from \cite{griffith2013policy}, and PACMAN without symbolic planner (AC with Human Feedback) as our ablation analysis. All plotting curves are averaged over 10 runs, and the shadow around the curve denotes the variance.

\begin{figure*}[htb!]
\begin{subfigure}{.33\textwidth}
\centering
    \includegraphics[height=4.2cm,width=4.4cm]{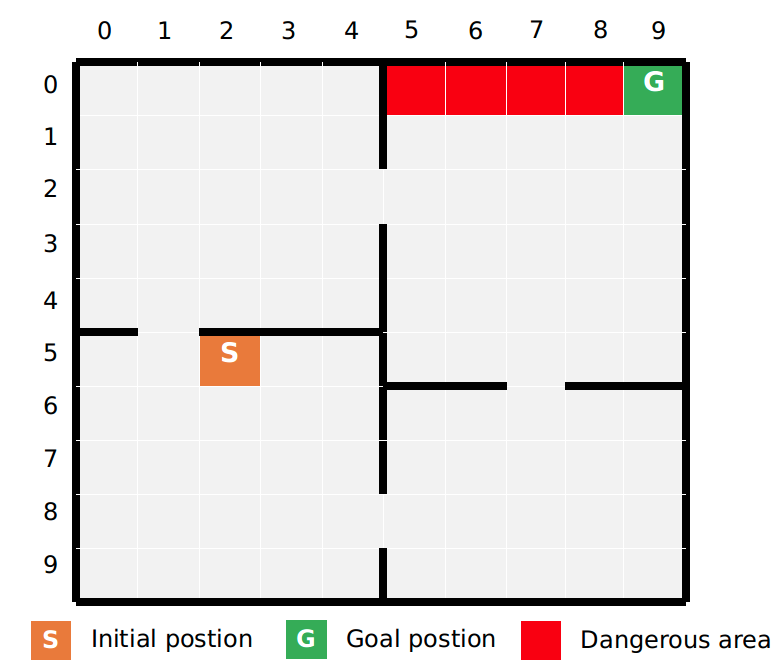}
  \subcaption{Four rooms domain}
  \label{fig:4room}
 \end{subfigure}
\begin{subfigure}{.33\textwidth}
\centering
    \includegraphics[height=4.2cm,width=4.2cm]{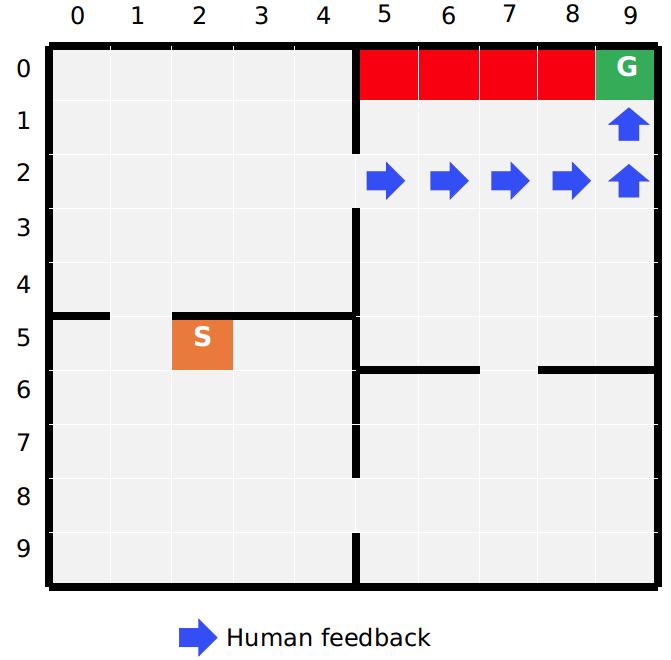}
  \subcaption{Helpful feedback}
  \label{fig:4room-help}
 \end{subfigure}
 \begin{subfigure}{.33\textwidth}
\centering
    \includegraphics[height=4.2cm,width=4.2cm]{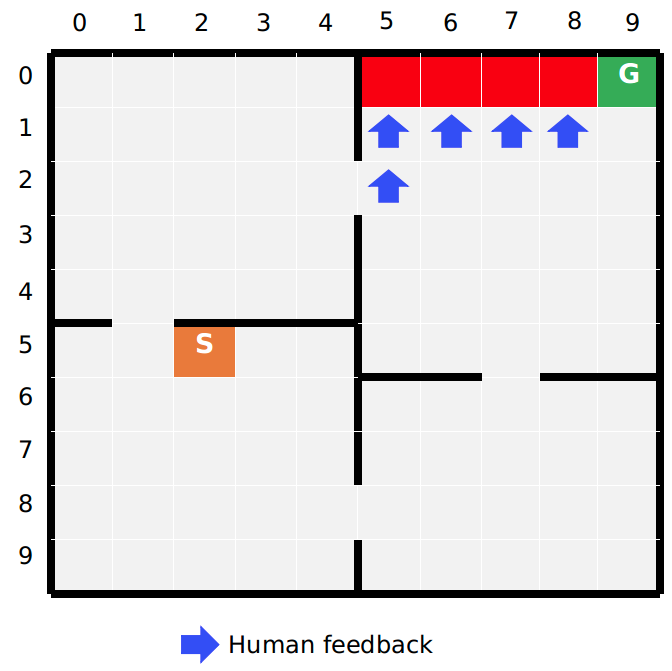}
  \subcaption{Misleading feedback}
  \label{fig:4room-mislead}
 \end{subfigure}
\caption{The snapshot of 2 scenarios on four rooms domain}
\label{fig:4room-2scens}
\end{figure*}

\begin{figure*}[htb!]
 \begin{subfigure}{.33\textwidth}
\centering
    \includegraphics[height=4.2cm,width=4.3cm]{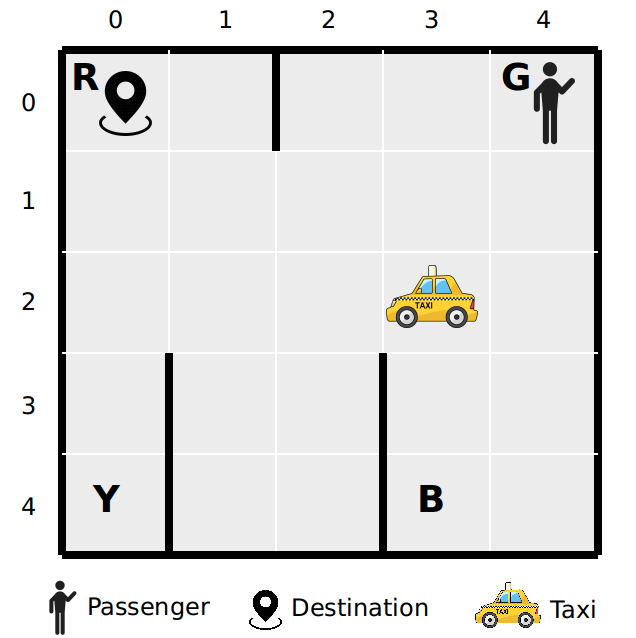}
  \subcaption{Taxi domain}
  \label{fig:taxi}
 \end{subfigure}
\begin{subfigure}{.33\textwidth}
\centering
    \includegraphics[height=4.2cm,width=4.2cm]{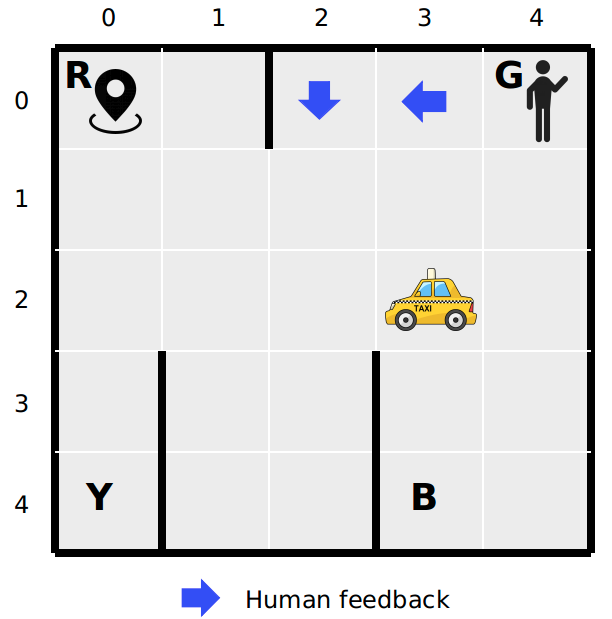}
  \subcaption{Helpful feedback}
  \label{fig:taxi-help}
 \end{subfigure}
 \begin{subfigure}{.33\textwidth}
\centering
    \includegraphics[height=4.2cm,width=4.2cm]{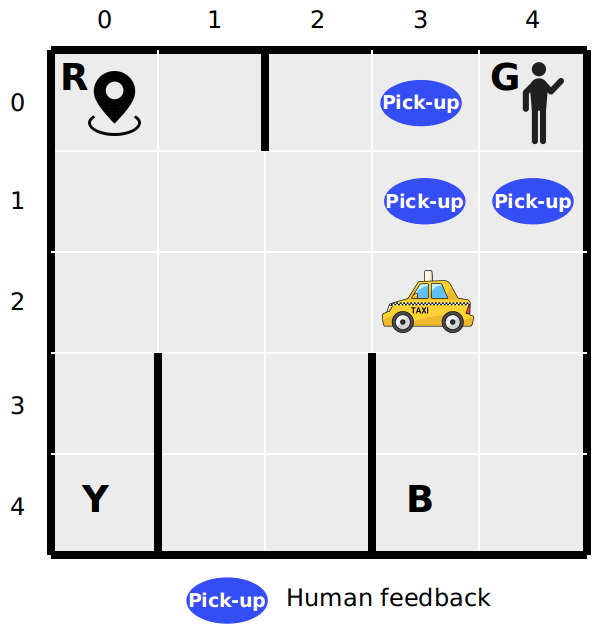}
  \subcaption{Misleading feedback}
  \label{fig:taxi-mislead}
 \end{subfigure}
\caption{The snapshot of 2 scenarios on taxi domain}
\label{fig:taxi-2scens}
\end{figure*}


\begin{figure*}[htb!]
\begin{subfigure}{.24\textwidth}
    \includegraphics[height=2.7cm,width=3.9cm]{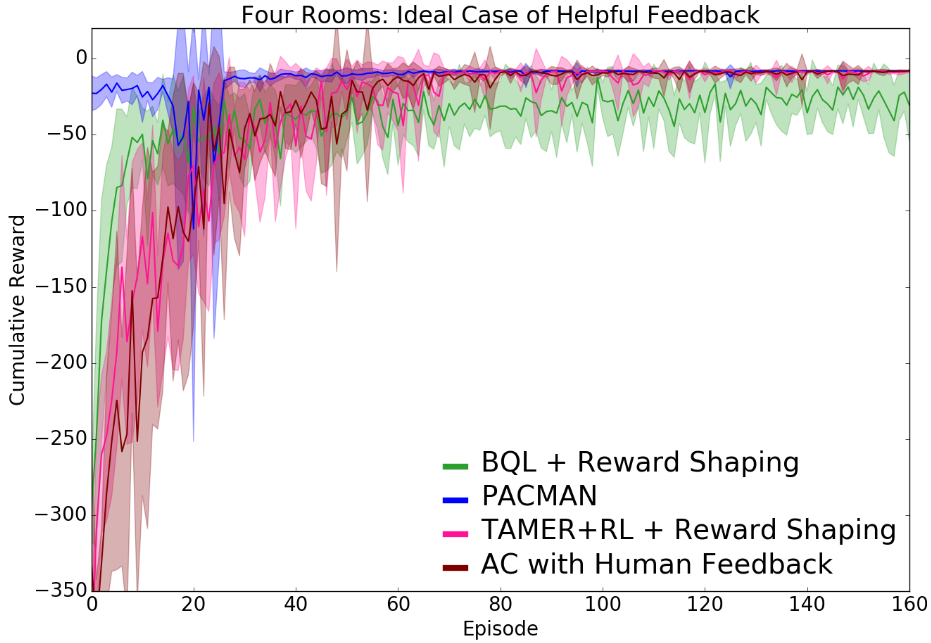}
  \subcaption{Ideal case}
  \label{fig:ideal-4room-help}
 \end{subfigure}
 \begin{subfigure}{.24\textwidth}
    \includegraphics[height=2.7cm,width=3.9cm]{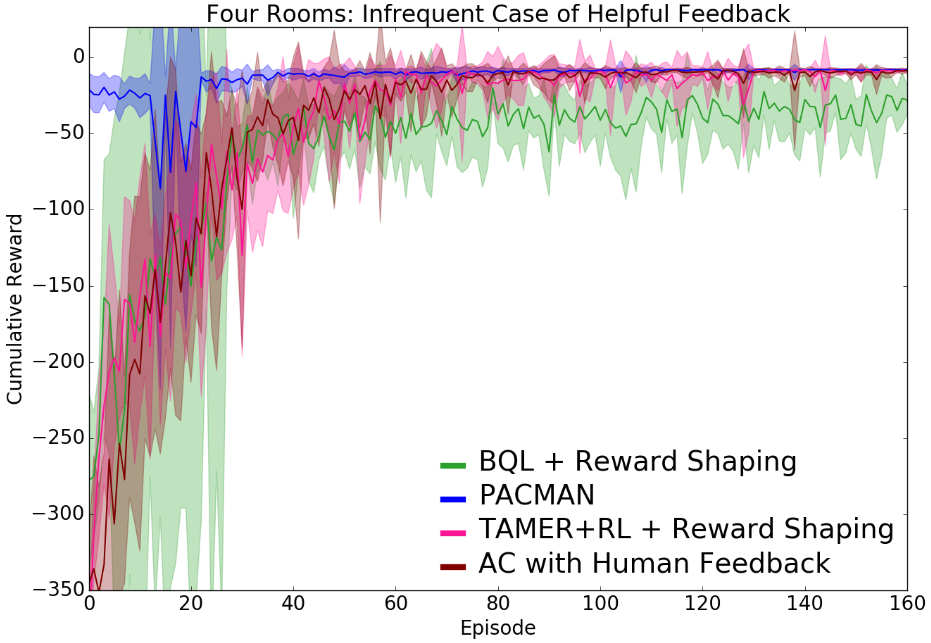}
  \subcaption{Infrequent case}
  \label{fig:infrequent-4room-help}
\end{subfigure}
 \begin{subfigure}{.24\textwidth}
    \includegraphics[height=2.7cm,width=3.9cm]{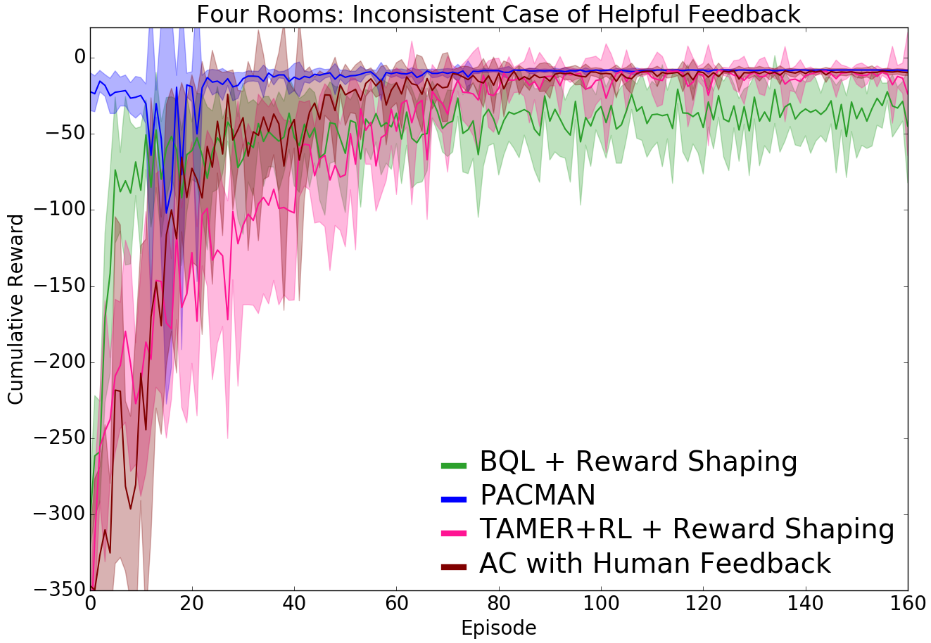}
  \subcaption{Inconsistent case}
 \label{fig:inconsistent-4room-help}
\end{subfigure}
 \begin{subfigure}{.24\textwidth}
    \includegraphics[height=2.7cm,width=3.9cm]{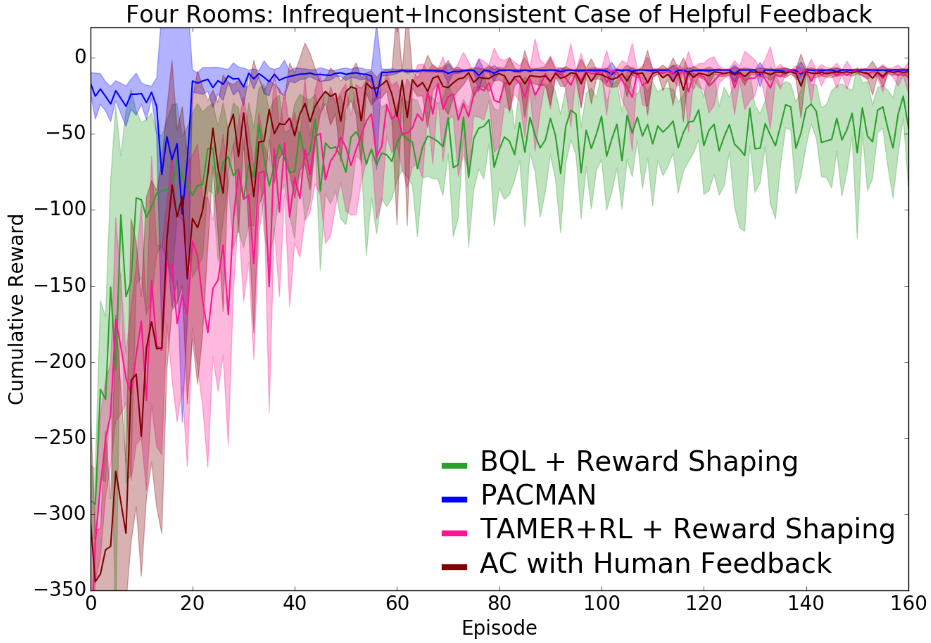}
  \subcaption{Infrequent+Inconsistent case}
  \label{fig:incons_infreq-4room-help}
\end{subfigure}
\caption{Four rooms with helpful feedback: learning curves}
\label{4roomhelpful}
\end{figure*}


\subsection{Four Rooms}
Four rooms domain is shown in Fig.~\ref{fig:4room}. In this 10$\times$10 grid, there are 4 rooms and an agent navigating from the initial position (5,2) to the goal position (0,9). If the agent can successfully achieve the task, it would receive a reward of +5. And it may obtain a reward of -10 if the agent steps into the red grids (dangerous area). Each move will cost -1. The human feedback of Four Rooms domain concerns 2 scenarios:


\begin{figure*}[htb!]
\begin{subfigure}{.24\textwidth}
    \includegraphics[height=2.7cm,width=3.9cm]{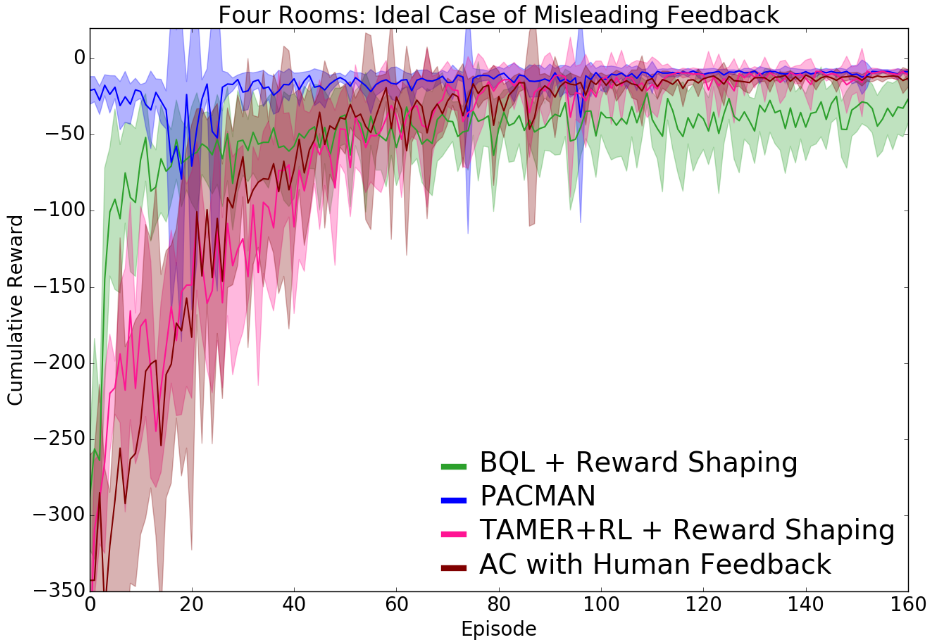}
  \subcaption{Ideal case}
  \label{fig:ideal-4room-mislead}
 \end{subfigure}
 \begin{subfigure}{.24\textwidth}
    \includegraphics[height=2.7cm,width=3.9cm]{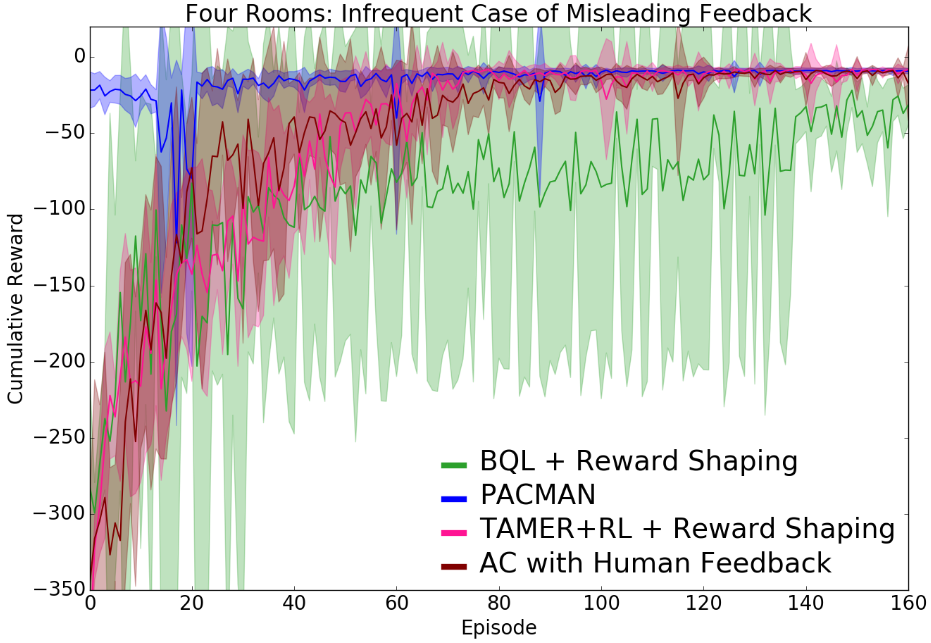}
  \subcaption{Infrequent case}
  \label{fig:infrequent-4room-mislead}
\end{subfigure}
 \begin{subfigure}{.24\textwidth}
    \includegraphics[height=2.7cm,width=3.9cm]{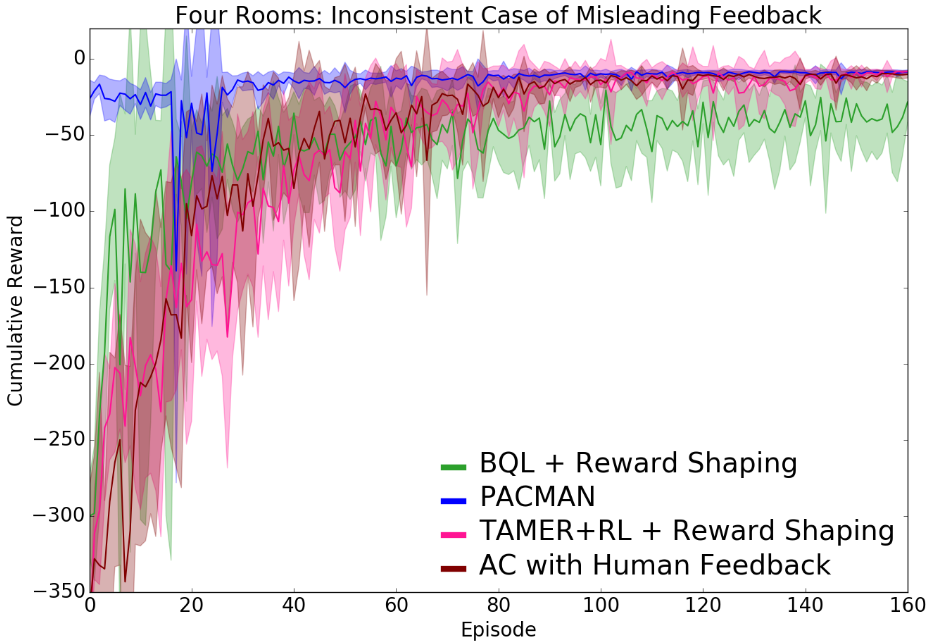}
  \subcaption{Inconsistent case}
  \label{fig:inconsistent-4room-mislead}
\end{subfigure}
 \begin{subfigure}{.24\textwidth}
    \includegraphics[height=2.7cm,width=3.9cm]{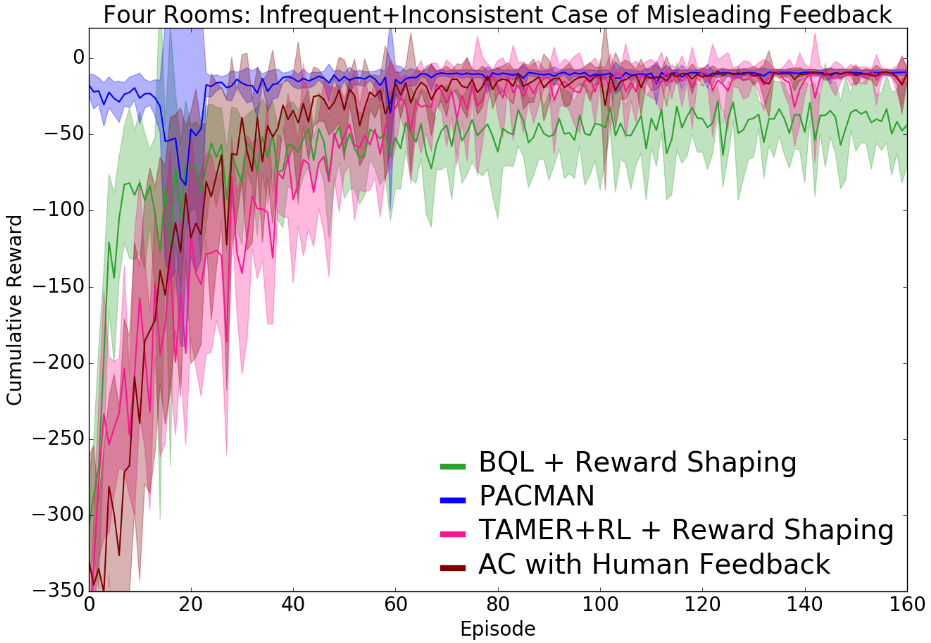}
  \subcaption{Infrequent+Inconsistent case}
  \label{fig:incons_infreq-4room-mislead}
\end{subfigure}
\caption{Four rooms with misleading feedback: learning curves}
\label{4roommislead}
\end{figure*}




\begin{figure*}[htb!]
\begin{subfigure}{.24\textwidth}
    \includegraphics[height=2.7cm,width=3.9cm]{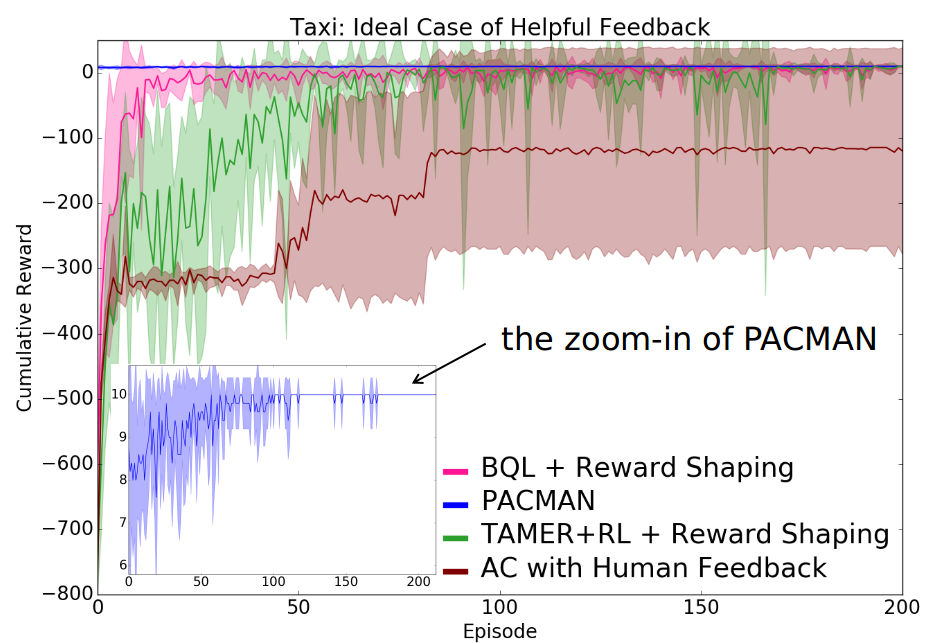}
  \subcaption{Ideal case}
  \label{fig:ideal-taxi-help}
 \end{subfigure}
 \begin{subfigure}{.24\textwidth}
    \includegraphics[height=2.7cm,width=3.9cm]{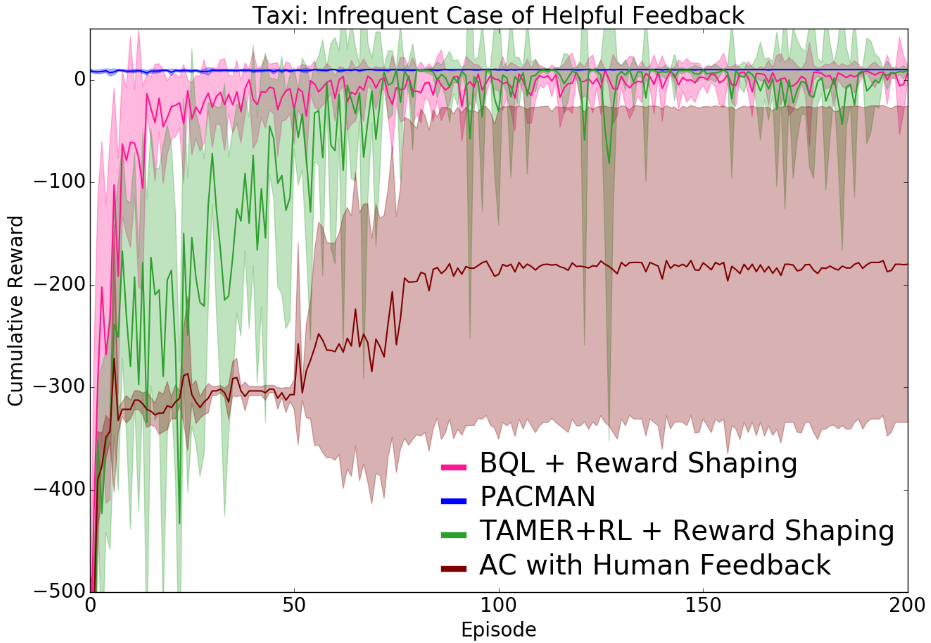}
  \subcaption{Infrequent case}
  \label{fig:infrequent-taxi-help}
\end{subfigure}
 \begin{subfigure}{.24\textwidth}
    \includegraphics[height=2.7cm,width=3.9cm]{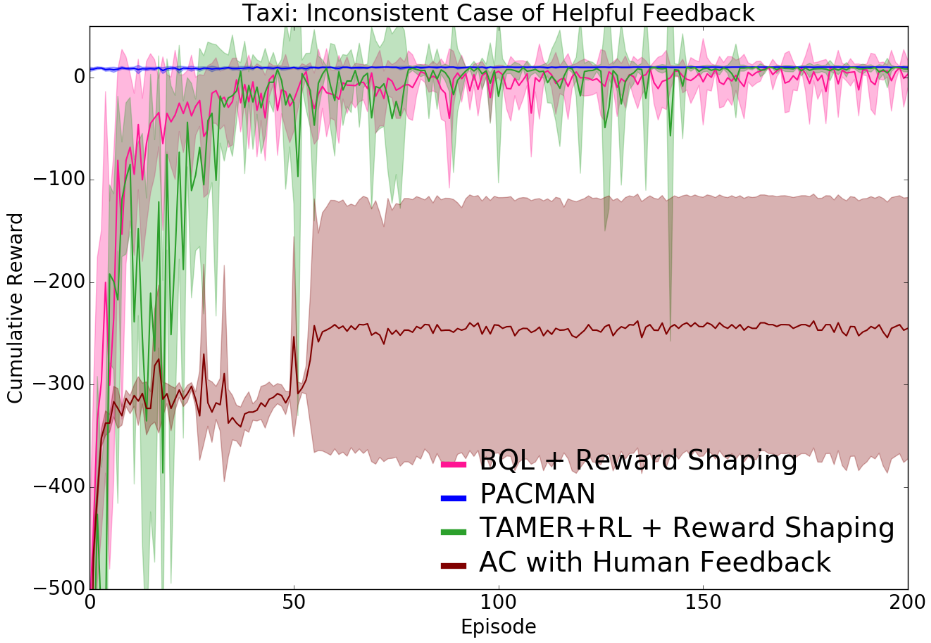}
  \subcaption{Inconsistent case}
  \label{fig:inconsistent-taxi-help}
\end{subfigure}
 \begin{subfigure}{.24\textwidth}
    \includegraphics[height=2.7cm,width=3.9cm]{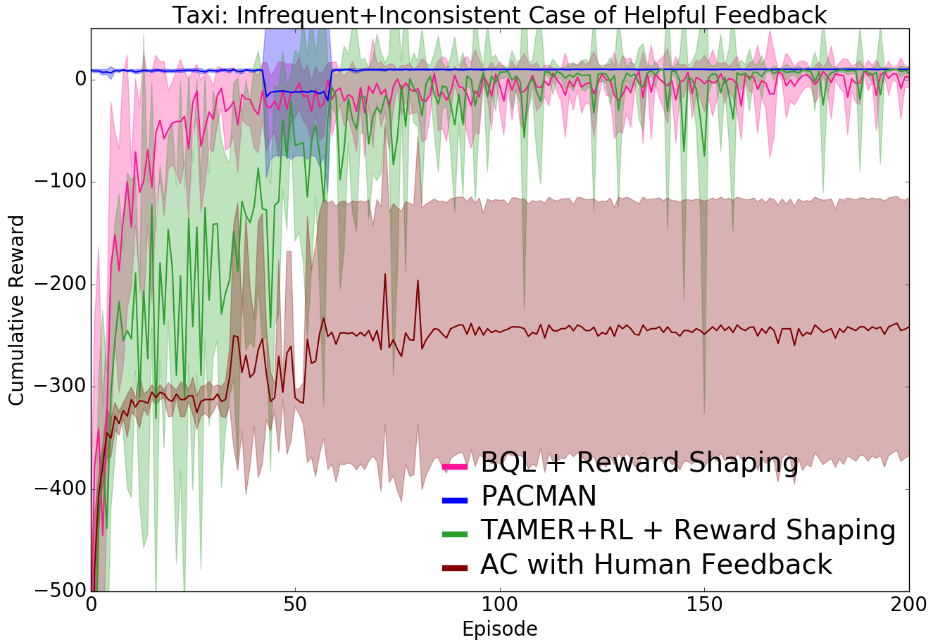}
  \subcaption{Infrequent+Inconsistent case}
  \label{fig:incons_infreq-taxi-help}
\end{subfigure}
\caption{Taxi with helpful feedback: learning curves}
\label{taxihelpful}
\end{figure*}


\begin{figure*}[htb!]
\begin{subfigure}{.24\textwidth}
    \includegraphics[height=2.7cm,width=3.9cm]{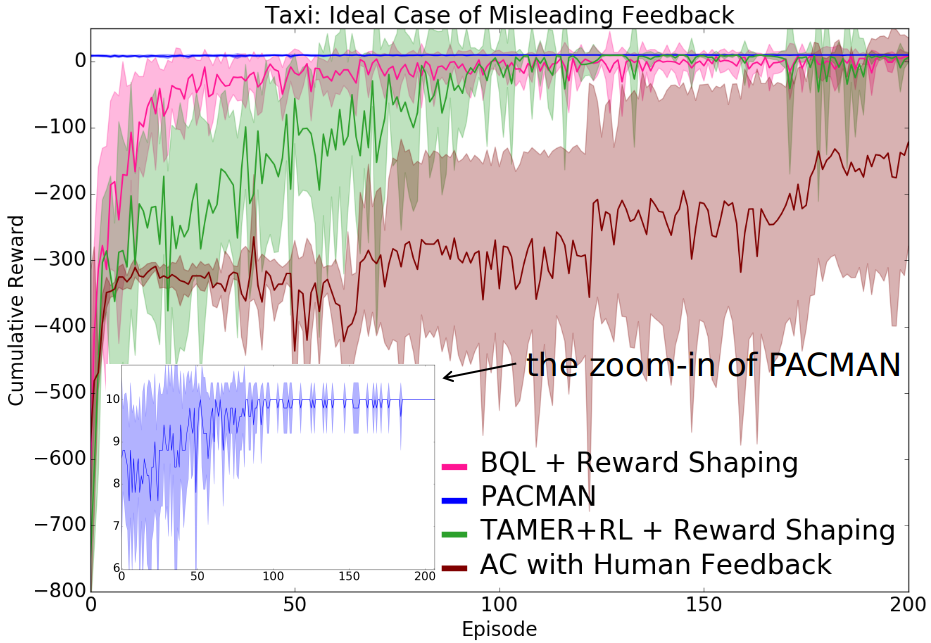}
  \subcaption{Ideal case}
  \label{fig:ideal-taxi-mislead}
 \end{subfigure}
 \begin{subfigure}{.24\textwidth}
    \includegraphics[height=2.7cm,width=3.9cm]{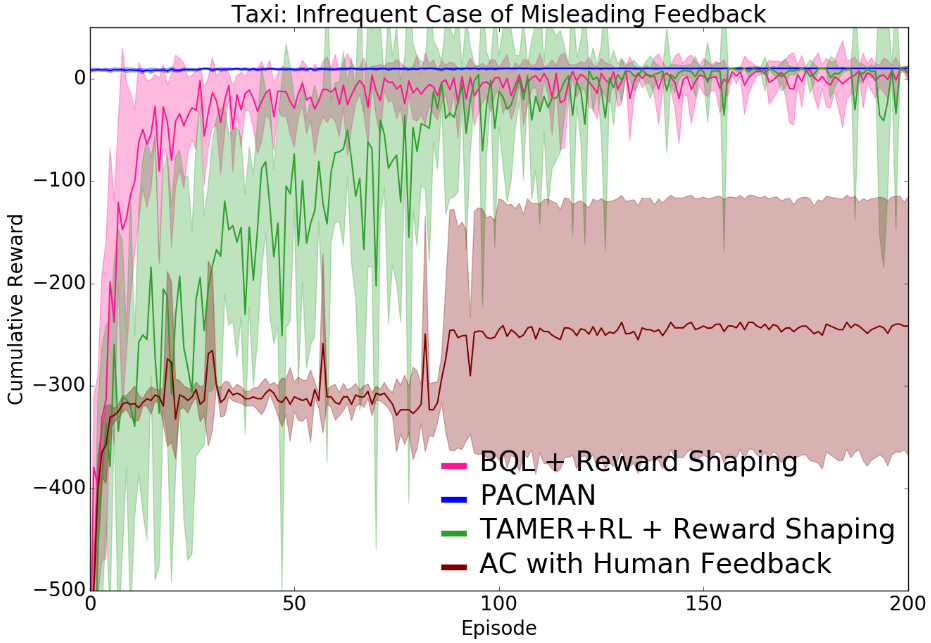}
  \subcaption{Infrequent case}
  \label{fig:infrequent-taxi-mislead}
\end{subfigure}
 \begin{subfigure}{.24\textwidth}
    \includegraphics[height=2.7cm,width=3.9cm]{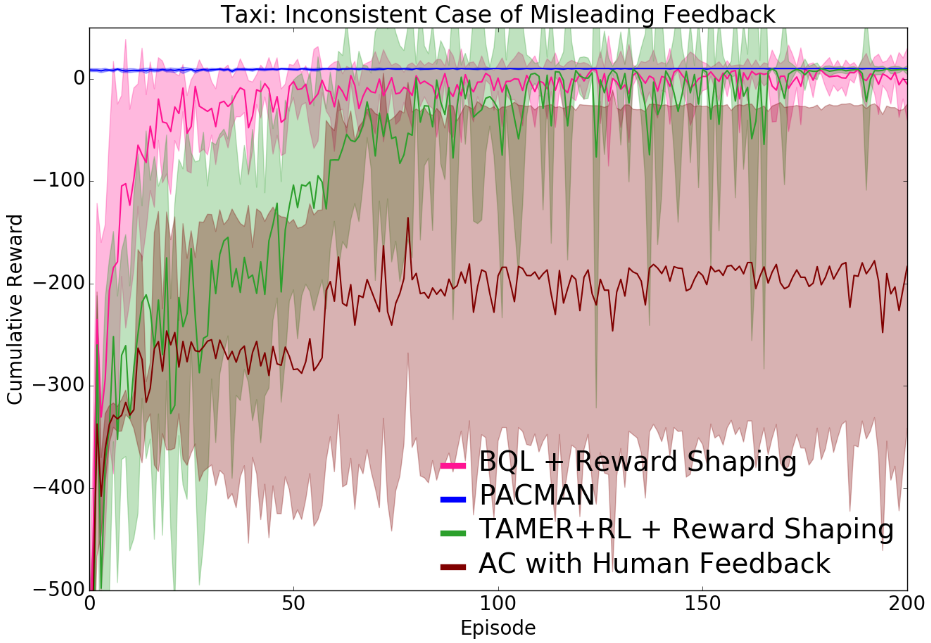}
  \subcaption{Inconsistent case}
  \label{fig:inconsistent-taxi-mislead}
\end{subfigure}
 \begin{subfigure}{.24\textwidth}
    \includegraphics[height=2.7cm,width=3.9cm]{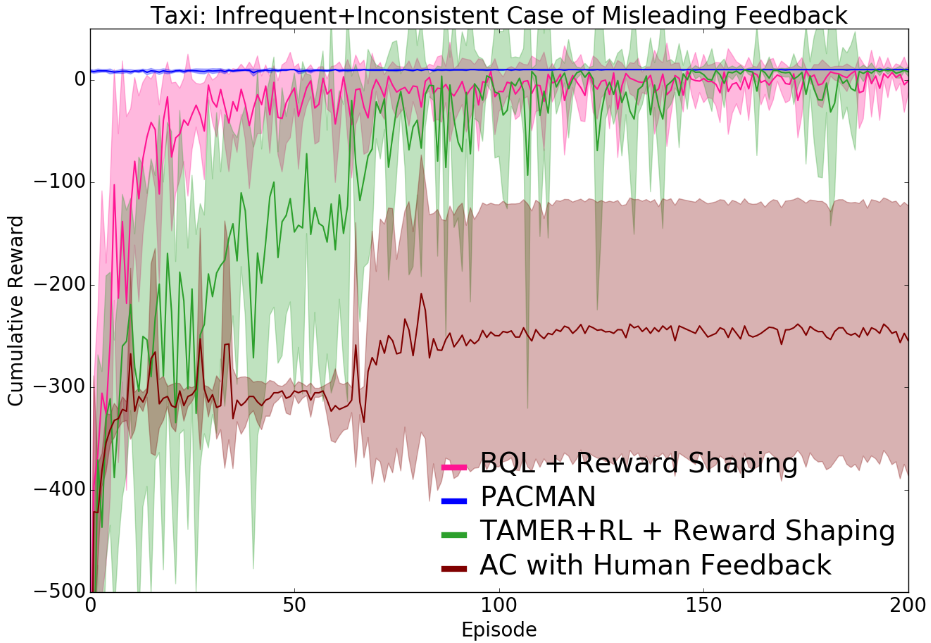}
  \subcaption{Infrequent+Inconsistent case}
  \label{fig:incons_infreq-taxi-mislead}
\end{subfigure}
\caption{Taxi with misleading feedback: learning curves}
\label{taximislead}
\end{figure*}


\begin{itemize}
\item Helpful feedback: consider an experienced user that wants to help the agent to navigate safer and better, such that the agent can stay away from the dangerous area and reach the goal position with the shortest path. Therefore, human feedback can guide the agent to improve its behavior towards the task, as shown in Fig.~\ref{fig:4room-help}.
\item Misleading feedback: consider an inexperienced user who doesn't know there is a dangerous area, but wants the agent to step into those red grids (Fig.~\ref{fig:4room-mislead}). In this case, human feedback contradicts with the behavior that the agent learns from an environmental reward.
\end{itemize}

The results are shown in Fig.~\ref{4roomhelpful} and Fig.~\ref{4roommislead}. Obviously, PACMAN has a jump-start and quickly converged with small variance,
compared to BQL Reward Shaping, TAMER+RL Reward Shaping, and AC with Human Feedback under four different cases. This is because symbolic planning leading to goal-directed behavior biasing exploration. Though the infrequent case, inconsistent case, and their combination case for both helpful feedback and misleading feedback can lead to more uncertainties, the performance of PACMAN remains unaffected, which means more robust than others.
Meticulous readers may find that there is a large variance in the initial stage of PACMAN, especially in Fig.~\ref{4roomhelpful}, Fig.~\ref{4roommislead}, this is due to the reason that the symbolic planner will first generate a short plan that is reasonably well, then the symbolic planner will perform exploration by generating longer plans. After doing the exploration, the symbolic planner will converge to the short plan with the optimal solution. But the large variance at the initial phase of PACMAN can be partially alleviated by setting the maximal number of actions in a plan to reduce plan space.

\subsection{Taxi Domain}
Taxi domain concerns a 5$\times$5 grid (Fig.~\ref{fig:taxi}) where a taxi needs to navigate to a passenger, pick up the passenger, then navigate to the destination and drop off the passenger. Each move has a reward of -1. Successful drop-off received a reward of +20, while improper pick-up or drop-off would receive a reward of -10. When formulating the domain symbolically, the precondition of performing picking up a passenger is specified that the taxi has to be located in the same place as the passenger. We consider human feedback in the following two scenarios:
\begin{itemize}
\item Helpful feedback: consider the rush hour, the passenger can suggest a path that would guide the taxi to detour and avoid the slow traffic, which is shown in Fig.~\ref{fig:taxi-help}. The agent should learn a more preferred route from human's feedback.
\item Misleading feedback: consider a passenger who is not familiar enough with the area and may inaccurately inform the taxi of his location before approaching the passenger (Fig.~\ref{fig:taxi-mislead}), which is the wrong action and will mislead the taxi. In this case, the feedback conflicts with symbolic knowledge specified by PACMAN and the agent should learn to ignore such feedback.
\end{itemize}

\noindent The results are shown in Fig.~\ref{taxihelpful} and Fig.~\ref{taximislead}. In the scenarios of both helpful and misleading feedback, the curve of PACMAN has the smallest variance so that it looks like a straight line, whereas it actually has the learning process (the zoom-in curve shown in the figures of the ideal case). But in the case of Infrequent+Inconsistent, there is a big chattering in the initial stage of PACMAN, that's because the symbolic planner is trying some longer plans to do the exploration. In the misleading feedback scenario, the learning speed of the other methods except for PACMAN is quite slow. That's because the human feedback will misguide the agent to perform the improper action that can result in the penalty, and the agent needs a long time to correct its behavior via learning from the environmental reward. But PACMAN keeps unaffected in this case due to the symbolic knowledge that a taxi can pick up the passenger only when it moves to the passenger's location.


\section{Conclusion}
In this paper we propose the PACMAN framework, which takes into consideration of the prior knowledge, learning from environmental rewards and human teaching together and jointly contribute to obtaining the optimal policy.
Experiments demonstrate that PACMAN tends to lead to a significant jump-start at early stages of learning, converge faster with reduced variance, and perform robustly to inconsistent, infrequent, and even misleading human feedback.
Our future work involves investigation of using PACMAN to perform decision-making from high-dimensional sensory input such as pixel images, autonomous driving where the vehicle can learn human's preference on comfort and driving behaviors, as well as multi-agent systems such as mobile service robots.

\section*{Acknowledgment}
This research was supported in part by the National Science Foundation (NSF) under grants NSF
IIS-1910794 and Amazon Research Award.

\bibliographystyle{aaai}
\bibliography{reference}

\end{document}